# Unsupervised Anomaly Detection and Localization with Generative Adversarial Networks

Khouloud Abdelli[(1)], Matteo Lonardi[(2)], Jurgen Gripp[(3)], Samuel Olsson[(3)],
Fabien Boitier[(4)], and Patricia Layec[(4)]

[(1)] Nokia Bell Labs, Germany Khouloud.Abdelli@nokia.com
[(2)] Nokia, Vimercate, Italy, [(3)] Nokia, Murray Hill, NJ, 07974 USA, [(4)] Nokia Bell Labs, France

**Abstract** *We propose a novel unsupervised anomaly detection approach using generative adversarial networks and SOP-derived spectrograms. Demonstrating remarkable efficacy, our method achieves over 97% accuracy on SOP datasets from both submarine and terrestrial fiber links, all achieved without the need for labelled data.* ©2024 The Author(s)

## Introduction

Monitoring the state of polarization (SOP) along optical fibers is pivotal for early detection of disturbances, indicating potential issues like fiber breakage [1-4]. However, conventional anomaly detection methods, such as threshold-based systems and rule-based approaches, struggle to accurately discern normal fluctuations from genuine anomalies, leading to false alarms or missed detections. Setting appropriate thresholds or rules is complex and may not cover the full spectrum of potential anomalies.

To overcome these limitations, integrating machine learning (ML) models holds promise for anomaly detection in optical networks. ML algorithms can learn complex patterns and relationships from labelled data, improving detection accuracy [5-6]. However, the scarcity of labelled data, particularly for rare anomalies, limits supervised ML approaches. Unsupervised ML methods provide a solution by autonomously learning the structure of normal data and detecting anomalies based on deviations, enhancing detection even with limited anomaly data [7-9].

In this paper, we present AnoGAN, an unsupervised anomaly detection method utilizing Generative Adversarial Networks (GANs) [10] and SOP-derived spectrograms. GANs learn a representation of normal behavior solely from normal training spectrograms. Trained GANs can only generate what is considered normal, thus unable to reconstruct possible abnormal regions in the spectrogram. We validate our approach using submarine and terrestrial link datasets.

## Anomaly Detection Framework

The overall framework, delineated in Fig. 1, encompasses two core modules: offline training and online anomaly detection.

Within the offline training module, we employ a pre-processed historical SOP dataset (the time series of the Stokes parameters) collected during normal operation to train the AnoGAN model. The Stokes time series are divided into non-overlapping sequences of fixed length $l$. These sequences are then transformed into time-frequency spectrograms using the short-time Fourier transform (STFT). These spectrograms are inputted into AnoGAN for training. The structure of AnoGAN, illustrated in Fig. 2, comprises two adversarial modules: a generator $G$ and a discriminator $D$. $G$ learns a distribution $p_g$ over data $x$ by mapping samples $z$, 1D vectors of uniformly distributed input noise sampled from the latent space $Z$, to 2D images (spectrograms) in the image space manifold $X$, which represents the variability of the training data (the purple region in Fig. 2 (b)). In this setup, $G$'s architecture resembles a convolutional decoder with strided convolutions. $D$, on the other hand, is a standard convolutional neural network that maps a 2D image (spectrogram) to a single scalar value, $D(.)$. The output $D(.)$ represents the probability that the input received by $D$ is a real spectrogram $x$ sampled from the training data $X$ or generated by $G$. $D$ and $G$ are concurrently optimized through a two-player minimax game with a value function $V(G,D)$ [10]

$$\min_G \max_D V(G,D) = \mathbb{E}_{x \sim p_{data}(x)}[\log D(x)] + \mathbb{E}_{z \sim p_z(z)}[\log(1 - D(G(z)))] \quad (1)$$

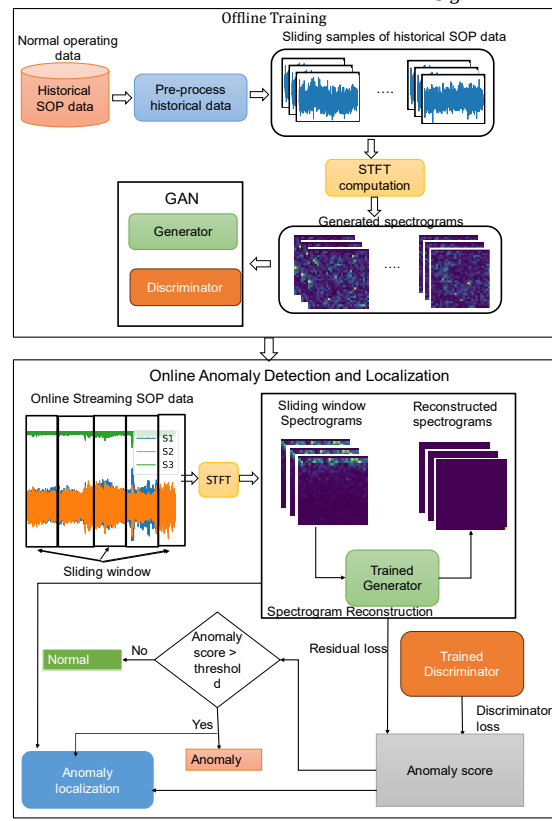

Fig. 1: Proposed framework for anomaly detection and localization in SOP derived spectrograms using GANs.

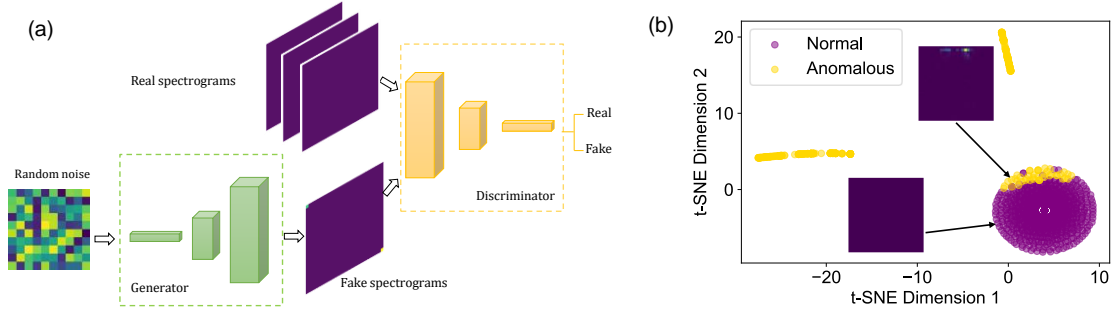

Fig. 2: (a) Deep convolutional generative adversarial network. (b) t-SNE embedding of normal (purple) and anomalous images (gold) from last convolution layer of the discriminator.

$D$ is trained to classify real examples as "real" and generated samples as "fake," while $G$ aims to deceive $D$. $G$ iteratively refines its generation process to produce realistic spectrograms, while $D$ enhances its discrimination skills to distinguish between real and generated ones.

The real-time anomaly detection module utilizes the trained AnoGAN model in conjunction with a snapshot of the streaming SOP data to calculate the corresponding anomaly score. The data reconstruction module first splits the streaming SOP data into sliding window samples, then converts them into spectrograms, which are finally reconstructed by the generator (see. Spectrogram Reconstruction Module in Fig. 1). The anomaly detection module calculates an anomaly score ($L$) by considering both the discriminator loss ($L_D$) and the residual loss ($L_R$) between the original and reconstructed spectrogram samples. $L_R$ represents the reconstruction error of the generated spectrogram sample, quantifying the visual dissimilarity between the sliding window spectrogram ($x$) and the reconstructed spectrogram ($G(z)$). It is computed as:

$$L_R(z) = \sum |x - G(z)| \quad (2)$$

A rich intermediate feature representation of the discriminator is adopted to compute $L_D$, leveraging $D$ as both a classifier and a feature extractor, rather than solely relying on its scalar output. $L_D$ is computed as [11]:

$$L_D(z) = \sum |f(x) - f(G(z))| \quad (3)$$

where $f(.)$ represents the output of an intermediate discriminator layer, capturing image (spectrogram) statistical properties. The overall anomaly score is defined as the weighted sum of both losses, expressed as:

$$L = \lambda L_R + (1 - \lambda) L_D \quad (4)$$

where $\lambda$ is a hyper-parameter between 0 and 1, which balances the weight between $L_R$ and $L_D$.

If the anomaly score of the current spectrogram sample surpasses a predefined threshold, the online anomaly detection algorithm flags an abnormal event. After anomaly detection, localization involves identifying abnormal events within the spectrogram both spectrally and temporally. This is done by computing residuals, representing pixel-wise differences between original and reconstructed spectrograms. Larger residuals indicate potential anomalies. Applying a threshold to these residuals identifies abnormal regions, with pixels exceeding the threshold marked as anomalies. Abnormal regions within spectrograms are visualized using color overlays.

**Validation on different datasets**

We validate our proposed approach using SOP datasets from various optical links. We first use the Curie Data [12] encompasses SOP recordings from the 10,000 km Curie submarine cable, connecting Los Angeles, California, and Valparaiso, Chile. It includes normal optical telecommunications traffic traces recorded from June 1 to July 12, 2020, as well as recordings of moderate and large earthquakes between December 15, 2019, and September 4, 2020. SOP recordings from normal operations are segmented into 4000-unit sliding windows, converted into spectrograms, and utilized for training the AnoGAN model. SOP recordings from earthquakes, along with a subset of normal operation samples not used in training, are employed for testing.

We then use "Terrestrial SOP Experiment Data (TSED)" dataset [13]. TSED includes SOP measurements from terrestrial links during normal operations and mechanical vibrations (anomalies), recorded using the experimental setup in Fig. 3. Initially, SOP measurements captured normal operation patterns without external factors like a robot arm or fan; only a polarization scrambler altered the initial polarization state. Anomalies were induced using a robot arm, executing movements like bending and shaking, and an IoT-controlled fan, causing disturbances like patch cord flapping. SOP recordings, including only normal operation, were segmented into 2000-size windows, transformed into spectrograms, and used for AnoGAN training. Testing involved both mechanical event samples (anomalies) and unseen normal samples.

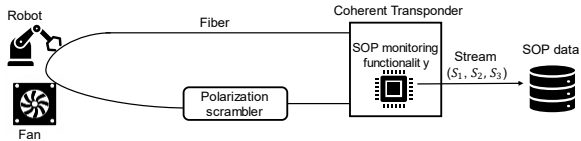

Fig. 3: Experimental setup for recording SOP data.

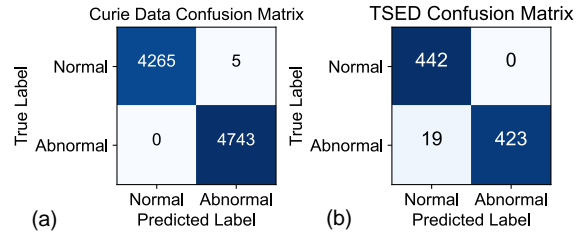

Fig. 4: Confusion matrices for a) Curie data and b) TSED.

**Results**

Our approach is evaluated on unseen test datasets containing both normal and abnormal instances. Performance is assessed using accuracy (Acc), precision (Pre), recall (Rec), F1 score, and area under the curve (AUC). Tab. 1 presents a detailed summary of the outcomes for various datasets under different Stokes parameters inputs. We explored the impact of training our model using all Stokes parameters, as well as with a single Stokes parameter. We consistently achieve robust performance across both datasets, with all metrics exceeding 97%. We note that performance is superior for submarine cables, which, as anticipated, tend to be less affected by noise and environmental disturbances compared to terrestrial links. The results also indicate that training our model with just one Stokes parameter yields comparable performance to that achieved when utilizing all Stokes parameters. This offers a potential reduction in computational complexity and resource demands.

Fig. 4 displays the confusion matrices, demonstrating precision in differentiating normal from abnormal instances. In Fig. 4(a), we see that all earthquakes are identified, including those of moderate amplitude. In Fig. 4(b), there are only rare misclassifications of some abnormal instances seen as normal. The t-SNE Embedding (Fig. 2(b)) demonstrates the discriminative ability of the final convolution layer features to distinguish between normal and anomalous images, showcasing the effectiveness of our AnoGAN in capturing meaningful variations in normal anatomy.

Fig. 5 illustrates the results of the anomaly localization. It includes three columns: the input spectrogram, the reconstructed spectrogram by the trained generator, and the localization of anomalies highlighted in red. For normal behavior, the reconstructed spectrogram closely resembles the input, confirming accuracy. In contrast, for abnormal samples, the discrepancies between the input and reconstructed spectrograms clearly mark the anomalous regions. This allows precise identification of the anomaly's spatial and temporal coordinates within the spectrogram, pinpointing when the anomaly occurs.

Tab. 1: Performance Evaluation of AnoGAN on submarine and terrestrial link SOP datasets.

| Input | Data | Acc (%) | Pre (%) | Rec (%) | F1 (%) | AUC (%) |
|---|---|---|---|---|---|---|
| $(S_1, S_2, S_3)$ | Curie | 99.9 | 99.8 | 100 | 99.9 | 99.9 |
|  | TSED | 97.8 | 100 | 95.7 | 97.8 | 97.8 |
| $S_1$ | Curie | 98.9 | 98 | 100 | 99 | 98.9 |
|  | TSED | 97.8 | 100 | 95.7 | 97.8 | 97.8 |
| $S_2$ | Curie | 99.9 | 99.8 | 100 | 99.9 | 99.9 |
|  | TSED | 97.8 | 100 | 95.7 | 97.8 | 97.8 |

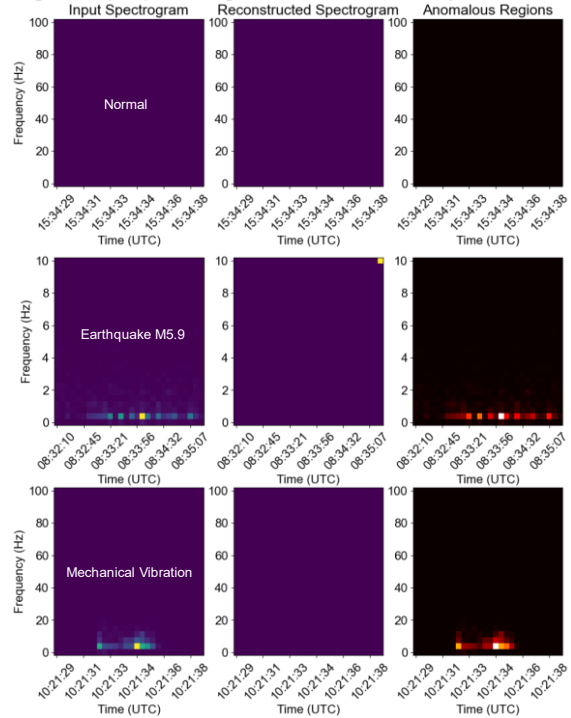

Fig. 5: Qualitive results of the localization of the anomalies using our approach.

**Conclusions**

We proposed a GAN-based anomaly detection method using SOP-derived spectrograms. Validated on submarine and terrestrial link datasets, our approach effectively distinguishes between normal and abnormal instances while accurately localizing anomalies within the spectrograms.